\documentclass[conference]{IEEEtran}
\IEEEoverridecommandlockouts
\usepackage{cite}
\usepackage{amsmath,amssymb,amsfonts}
\usepackage{algorithmic}
\usepackage{graphicx}
\usepackage{textcomp}
\usepackage{xcolor}
\usepackage{longtable}
\usepackage{multirow}

\usepackage[margin=1in]{geometry}
\usepackage{stfloats}

\usepackage{capt-of} 
\usepackage{cuted} 

\usepackage{colortbl} 
\usepackage{booktabs}
\usepackage{makecell}

\usepackage{hyperref}

\usepackage{authblk}

\def\BibTeX{{\rm B\kern-.05em{\sc i\kern-.025em b}\kern-.08em
    T\kern-.1667em\lower.7ex\hbox{E}\kern-.125emX}}
\begin{document}

\title{Face-MakeUp: Multimodal Facial Prompts for Text-to-Image Generation}

\author{Dawei Dai, Mingming Jia, Yinxiu Zhou, Hang Xing, Chenghang Li} 

\affil[]{Chongqing Key Laboratory of Computational Intelligence, Chongqing University of Posts \\ and Telecommunications, 400065, Chongqing, China.}

\maketitle

    \begin{strip} 
        \centering
        \includegraphics[width=0.96\textwidth]{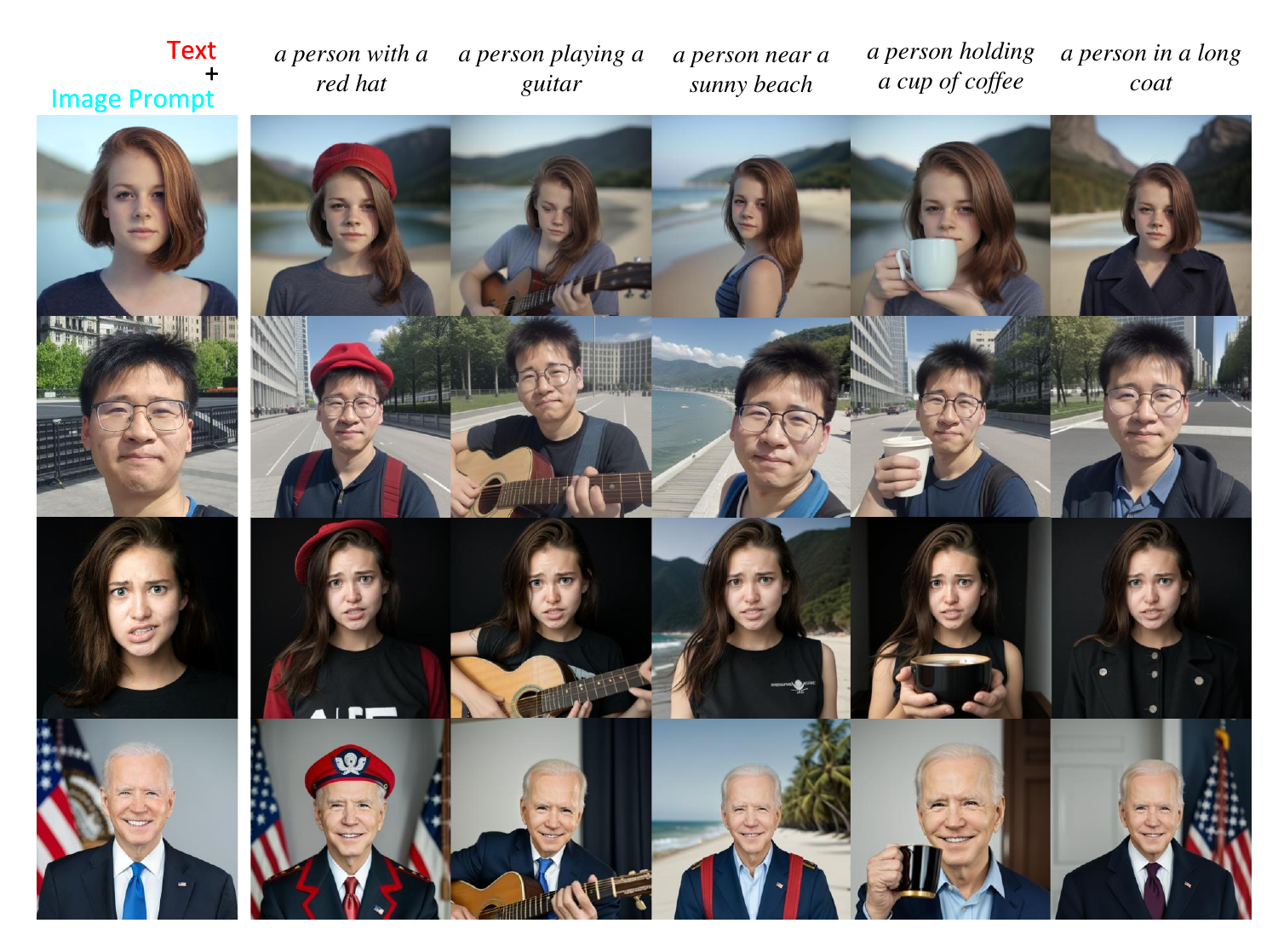}
        \captionof{figure}{Illustrations of our Face-MakeUp. The first column is the image prompts (reference), and others show the images generated by our model using reference images.}
        \label{fig:cover}
    \end{strip}

\begin{abstract} Facial images have extensive practical applications. Although the current large-scale text-image diffusion models exhibit strong generation capabilities, it is challenging to generate the desired facial images using only text prompt. Image prompts are a logical choice. However, current methods of this type generally focus on general domain. In this paper, we aim to optimize image makeup techniques to generate the desired facial images. Specifically, (1) we built a dataset of 4 million high-quality face image-text pairs (FaceCaptionHQ-4M) based on LAION-Face to train our Face-MakeUp model; (2) to maintain consistency with the reference facial image, we extract/learn multi-scale content features and pose features for the facial image, integrating these into the diffusion model to enhance the preservation of facial identity features for diffusion models. Validation on two face-related test datasets demonstrates that our Face-MakeUp can achieve the best comprehensive performance. All codes are available at: \href{https://github.com/ddw2AIGROUP2CQUPT/Face-MakeUp}{Face-MakeUp}  \end{abstract}

\section{Introduction}
\label{sec:intro}

Image generation has made remarkable strides with the success of recent large text-to-image diffusion models like DALL-E 2\cite{ramesh2022hierarchical}, Imagen\cite{saharia2022photorealistic}, and Stable Diffusion (SD)\cite{rombach2022high}. Stable Diffusion, by performing diffusion in latent image space, not only revolutionized the generation process but also significantly reduced computational costs, making it one of the dominant methods in recent years. Not only do they exhibit a remarkable capacity for generating high-quality, intricate, and imaginative images, but their practical applications in fields such as artistic creation, advertising design, and interactive entertainment have proven that they can significantly enhance work efficiency and open up new possibilities for innovation. \par

In practical applications, facial image generation have broad application across various aspects \cite{wei2023elite}, \cite{ju2023humansd}. However, crafting effective text prompts to produce desired face image can be challenging. Two fundamental reasons can be summarized as: (1) text alone may not adequately convey complex scenes or concepts. It is difficult for people to describe highly complex geometric and textural information using text, resulting in generated images that deviate from expectations. (2) The absence of large-scale image-text pair datasets with sufficiently detailed text results in models having insufficient capability to generate detailed images.\par

Considering these limitations of text prompts, image prompts are a logical choice, as they can convey more details and nuances than text. For example, IP-Adapter\cite{ye2023ip} extracts image features through CLIP\cite{radford2021learning} and fuse the multimodal information using cross-attention modules to guide the diffusion process, generating images that align with the reference image. Photomaker\cite{li2024photomaker} combines facial embeddings that projected into the CLIP space with text tokens to produce joint embeddings as a conditional input to guide the diffusion process. InstantID\cite{wang2024instantid} introduces an additional ID \& Landmark ControlNet\cite{zhang2023adding}, further improving the model's control efficiency, while it significantly boosts ID similarity, it sacrifices some editing capability and flexibility. PuLID\cite{guo2024pulid} combines contrastive loss and precise ID loss to effectively minimize interference with the reference while ensuring a high level of ID fidelity.\par

Although researchers have conducted a series of important studies and achieved certain successes in image-prompt text-to-image generation, there is a significant need for specialized optimization for facial image generation due to its wide range of applications. First, we constructed a large-scale facial image-text dataset (FaceCaptionHQ-4M) to train our dedicated facial model (Face-MakeUp). Second, we utilized both general and specialized facial visual encoders to extract multi-scale features from facial regions, and also employed a structure extractor to learn the structural (pose) information, integrating them into the pretained diffusion model, ensuring consistency between the facial regions of the generated images and the references. Experiments demonstrate that our Face-MakeUp achieves the best competitive outcomes across a broad range of evaluation metrics. Our contributions can be summarized as:\par

(1) \textbf{FaceCaptionHQ-4M.} For facial image generation task, we constructed a large-scale and high-quality facial image-text dataset, where text describe facial features.\par

(2) \textbf{Face-MakeUp Model.} We developed and optimized the image-prompt Text-to-Image diffusion model for facial image generation, and achieve the best consistency of facial image.\par

(3) \textbf{Open-source.} To facilitate research, we will release the following assets to the public: All the image-text pairs data, the model checkpoints, and the codebase for model training.\par

\begin{figure*}[htbp]
\centering
\includegraphics[width=\linewidth, trim=20px 20px 20px 20px, clip]{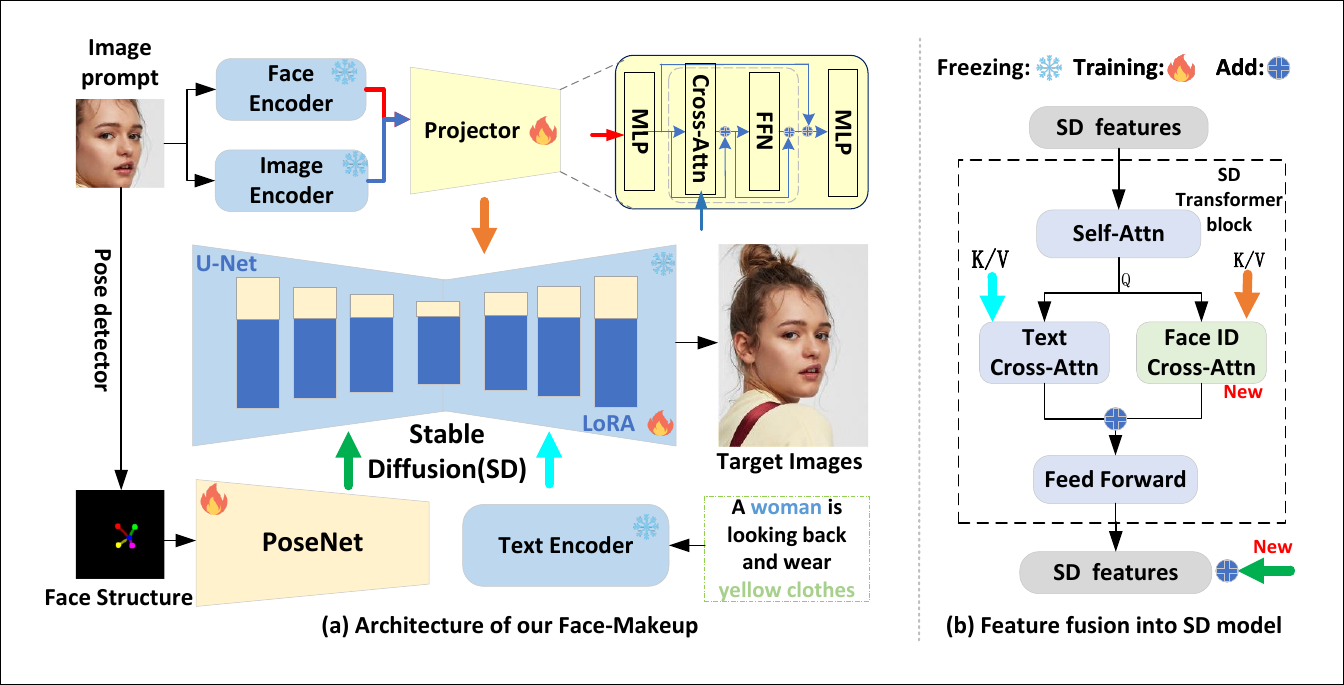}
\caption{Overview of our Face-MakeUp.}
\label{fig:overview}
\end{figure*}


\section{Method}
\subsection{Constructing FaceCaptionHQ-4M}
LAION-Face\cite{zheng2022general} is a large-scale facial image-text dataset. However, the quality of facial images in this dataset varies significantly. In this study, we cleaned the LAION-Face to construct a high-quality facial image-text data for image generation task. We utilize the information of FaceCaption-15M\cite{dai202415m} (each image in FaceCaption-15M corresponds one image in LAION-Face) to clean the LAION-Face data efficiently. Specifically: (1) We sorted the images in FaceCaption-15M by resolution, and selected the top 10M images from LAION-Face; (2) We removed black-and-white images by checking whether the mean of the standard deviation exceeds a set threshold to ensure the selection of color images; (3) We employed an OCR text detection model \cite{lee2012improving} to eliminate images containing a large amount of text; (4) We removed the group photos containing multiple faces by using the yolov5-face model\cite{qi2022yolo5face}; (5) We eliminated cartoon-style images using a cascade classifier based on LBP\cite{shetty2021facial} to detect anime-style faces within the images. Finally, we obtained a 4.2M high-quality human-scene images.

After finishing the selection of candidate images, we further processed the human photos. Specifically, (1) we proportionally expanded the width and height of the face region (as indicated by the bounding box that provided in FaceCaption-15M) by 1.5 times outward to encompass the upper body. We then cropped facial region into a square based on the smaller one (width or height); (2) we used Qwen2-VL\cite{wang2024qwen2} to generate more concise short text descriptions for the image.

\subsection{Architecture of Face-MakeUp}

Our purpose is to generate the image that aligns with the input text while maintaining a high degree of consistency with the reference image. In this study, we integrate fine-grained features of face into the feature space of the diffusion model (SD) while preserving posture, ensuring that the generated image reflects the identity characteristics and overall facial morphology of the reference portrait. 

Specifically, (1) the inputs of Face-MakeUp include a reference facial image, a pose map that extracted from the reference image, and text prompt; (2) facial features extraction modules, which includes general and specialized visual encoders as well as a learning module for pose map; (3) a pre-trained text-to-image diffusion model; and (4) a cross-attention module is designed to learn the joint representation of facial image (reference) and text prompts. In addition, embeddings of pose map are integrated through an additive way (Fig.~\ref{fig:overview}(b)). This final embeddings are then incorporated into the feature space of the diffusion model through an overlay method, which enriches the feature space of the diffusion model with more information of the reference facial image, thereby ensuring consistency between the generatacial image and the reference image.

\subsection{Extracting Representations for Facial Image-Text}

We need to extract or learn the embeddings of the prompt image and text. We utilize CLIP's Image-Encoder and Text-Encoder as general encoders to extract the representations of facial images and their corresponding texts, respectively. This is not enough. To enhance the fine-grained facial features and individual identity characteristics, we introduce ArcFace model \cite{deng2019arcface} as FaceEncoder. ArcFace is designed specifically for facial identity verification, generating highly discriminative and consistent facial embeddings in high-dimensional feature space. We combine the facial features extracted by ArcFace and CLIP's visual encoder through a classical cross-attention module (projector), thereby enhancing the representation capability of facial images.

\subsection{Learning Representations of Facial Pose}

To precisely maintain the structure consistency between the generated facial image and the prompt image, we incorporate facial pose information into the feature space of the text-to-image diffusion model. Specifically: (1) We first employ a pose detector\cite{deepinsight2020insightface} to obtain the pose map for the reference facial image; (2) We design a learnable module, named PoseNet, to learn the representations (embeddings) of the facial pose. To ensure that the feature vectors learned by this module can be seamlessly integrated into the corresponding layers of the diffusion model, the architecture of PoseNet is kept identical to that of the diffusion model.

\subsection{Fusion Strategy}

Our Face-MakeUp framework incorporates three additional embeddings into the diffusion model to enhance the consistency between the generated image and the reference. As shown in Fig.~\ref{fig:overview}(b): (1) We introduce a new cross-attention module (Face-ID Cross-Attn) in the diffusion model to fuse the visual features of the reference facial image with the feature space of the diffusion model; (2) The fused features of the reference facial image are further combined with the text features via addition, and then passed through a feedforward layer to achieve the feature fusion between the reference and the text; (3) the pose features of the reference are added to (2), enabling multi-modal features fusion of the face. Using this approach, additional modalities of features can also be integrated, allowing for even more precise facial image generation.

\begin{table*}[htbp]
    \centering
    \caption{\textbf{Comparisons with Existing Methods on Unsplash-face  and FaceCaption (Ours).}}
    \resizebox{\textwidth}{!} {%
    \Large
    \begin{tabular}{c@{\hspace{0.5cm}} l c c c c c c c}
        \toprule[2pt]
    
        &  & \textbf{CLIP-T} \(\uparrow\)  & \textbf{CLIP-I} \(\uparrow\)  & \textbf{DINO} \(\uparrow\)  & \textbf{FaceSim} \(\uparrow\) & \textbf{FID} \(\downarrow\) & \textbf{Attr\_c}  \(\uparrow\) & \textbf{VLM-score} \(\uparrow\) \\
        \hline
    
        \multirow{5}{*}{\rotatebox{90}{\makecell[c]{\textbf{Unsplash-Face}}}} 
         & \textbf{Ip-Adapter.}(2023)\cite{ye2023ip} & 27.7 & 64.9 & 37.6 & 53.2 & 226.9 & 3.0 & 65.3 \\
         & \textbf{PhotoMaker.}(2024)\cite{li2024photomaker} & 28.2 & 56.5 & 26.2 & 20.7 & 224.4 & 2.2 & 60.1 \\
         & \textbf{InstantID.}(2024)\cite{wang2024instantid} & 24.8 & 78.0 & 49.4 & \textbf{71.2} & 178.7 & 3.8 & 54.8 \\
         & \textbf{Pulid.}(2024)\cite{guo2024pulid} & \textbf{29.3} & 46.3 & 21.0 & 24.3 & 284.5 & 2.4 & 36.5 \\
         & \cellcolor{gray!20}\textbf{Ours} 
            & \cellcolor{gray!20}22.3
            & \cellcolor{gray!20}\textbf{82.1} 
            & \cellcolor{gray!20}\textbf{73.2} 
            & \cellcolor{gray!20}69.2 
            & \cellcolor{gray!20}\textbf{130.1} 
            & \cellcolor{gray!20}\textbf{4.0} 
            & \cellcolor{gray!20}\textbf{79.6} \\
        \hline
        
        \multirow{5}{*}{\rotatebox{90}{\makecell[c]{\textbf{FaceCaption} }}} 
         & \textbf{Ip-Adapter.}(2023)\cite{ye2023ip} & 26.78 & 69.7 & 48.0 & 59.2 & 195.4 & 3.2 & 63.2 \\
         & \textbf{PhotoMaker.}(2024)\cite{li2024photomaker} & 28.12 & 50.5 & 25.9 & 22.1 & 237.6 & 2.2 & 54.5 \\
         & \textbf{InstantID.}(2024)\cite{wang2024instantid} & 24.29 & 67.2 & 50.1 & 75.5 & 166.5 & 5.3 & 53.7 \\
         & \textbf{Pulid.}(2024)\cite{guo2024pulid} & \textbf{29.21} & 36.2 & 13.2 & 22.8 & 298.5 & 2.1 & 43.5 \\
         & \cellcolor{gray!20}\textbf{Ours} 
            & \cellcolor{gray!20}21.96
            & \cellcolor{gray!20}\textbf{87.4} 
            & \cellcolor{gray!20}\textbf{79.4} 
            & \cellcolor{gray!20}\textbf{77.8} 
            & \cellcolor{gray!20}\textbf{95.4} 
            & \cellcolor{gray!20}\textbf{6.3} 
            & \cellcolor{gray!20}\textbf{73.1} \\

        \toprule[2pt]
    \end{tabular}
    }
    \label{tab:metrics}
\end{table*}

\section{Experiments}

\subsection{Implementation Details}

\textbf{Training process.} We adopt stable-diffusion-v1-5\cite{rombach2022high} as our pre-trained Text-to-Image model. Correspondingly, the training data resolution is adjusted to 512x512. We use CLIP Vit-H/14\cite{schuhmann2022laion} and buffalo\_l \cite{deepinsight2020insightface} models as our Image-Encoder and Face-Encoder, respectively. The overall framework is trained using the Adam optimizer, with the learning rate set to 1e-4, over 500,000 steps on an 8*H800 GPU (94G), where the batch size per GPU is 10. Additionally, we apply 50\% random dropout to the image features extracted by CLIP. During inference, we use DDIM\cite{song2020denoising} as the sampler, with the number of steps set to 50 and the guidance scale set to 7.5.

\textbf{Metric.} (1) Some popular metrics, such as \textbf{DINO}\cite{caron2021emerging}, \textbf{CLIP-I}\cite{gal2022image}, \textbf{CLIP-T} and \textbf{FID}\cite{heusel2017gans} are commonly employed to evaluate the generated image. (2) We also employ \textbf{FaceSim} \cite{schroff2015facenet} to further evaluate facial similarity. (3) To score the realism, we introduce Qwen2VL\cite{wang2024qwen2} to evaluate the models \textbf{VLM-score}. In addition, we employ a detection model \cite{he2017adaptively} to recognize the attributes of the generated facial image, and the total count of detected attributes (\textbf{Attr\_c}) is used as a basis for fine-grained evaluation of models.	


\textbf{Datasets.} We evaluate the models on two dataset, in which each dataset contain 50 reference images. We design 20 different text prompts for each reference image. The first dataset is the Unsplash-Face selected from the People category of the Unsplash\footnote{https://unsplash.com} website, collected in October. The second one comes from the test set of our FaceCaptionHQ-4M. By modifying the prompt texts (as shown in Table~\ref{tab:prompt}), each model under evaluation generate 20 images for each reference image, resulting in a total of 1,000 image-text pairs per test dataset.

\begin{table}[htbp]
     
    \caption{Text prompts categorized by Clothing \& Accessories, Background, Action, and Style.}
    \scalebox{0.88}{
    
    \begin{tabular}{ll}    
        \toprule[1pt]

    \textbf{Category} & \textbf{Prompt} \\ 
    \hline  
    & \normalsize a \texttt{<class word>} with a red hat. \\
    \normalsize Clothing \& & \normalsize a \texttt{<class word>} in a long coat. \\ 
    \normalsize Accessory& \normalsize a \texttt{<class word>} wearing glasses. \\ 
    & \normalsize \textbf{...} \\
    \hline 
    & \normalsize a \texttt{<class word>} standing in a park. \\
    & \normalsize a \texttt{<class word>} in a cozy room. \\  
    \normalsize Background& \normalsize a \texttt{<class word>} near a sunny beach. \\
    & \normalsize \textbf{...} \\ 
    \hline 
    & \normalsize a \texttt{<class word>} running on a track. \\ 
    & \normalsize a \texttt{<class word>} holding a cup of coffee. \\ 
    \normalsize Action& \normalsize a \texttt{<class word>} playing a guitar. \\ 
    & \normalsize \textbf{...} \\  
    \hline 
    & \normalsize a \texttt{<class word>} dressed in formal attire. \\ 
    & \normalsize a \texttt{<class word>} wearing sportswear. \\ 
    \normalsize Outfit Style & \normalsize a \texttt{<class word>} dressed for a wedding. \\ 
    & \normalsize \textbf{...} \\ 
    \toprule[1pt]

    \end{tabular}
    \label{tab:prompt}
    }
    \end{table}
    
\subsection{Comparisons with Existing Methods}

We present the comparisons in Table~\ref{tab:metrics} and Fig.~\ref{fig:cover}. We can make the main observations as follows:

(1) In terms of the realism for generated facial images (VLM-score), our proposed Face-MakeUp significantly outperforms other models, indicating that our model can generate more realistic facial images. This is also demonstrated by the examples shown in Fig.~\ref{fig:cover}.

(2) Regarding attribute prediction in generated facial images (Attr\_{}c), facial images generated by Face-MakeUp contain more attributes than that of others, indicating that our model is capable of generating facial images that contain more fine-grained features.

(3) In terms of similarity between generated facial images and reference (CLIP-I, DINO, FaceSim, and FID), attributed to the diversified facial feature fusion mechanism, our model achieved seven first-place and one second-place performances across two test datasets.

(4) In terms of image-text similarity, our model is slightly lower than other models, mainly because the image contains not only faces but also other content. We mainly focus on optimizing the face region.

\subsection{Manual evaluation}

\begin{figure}[htbp]
\centering
\includegraphics[scale=0.43]{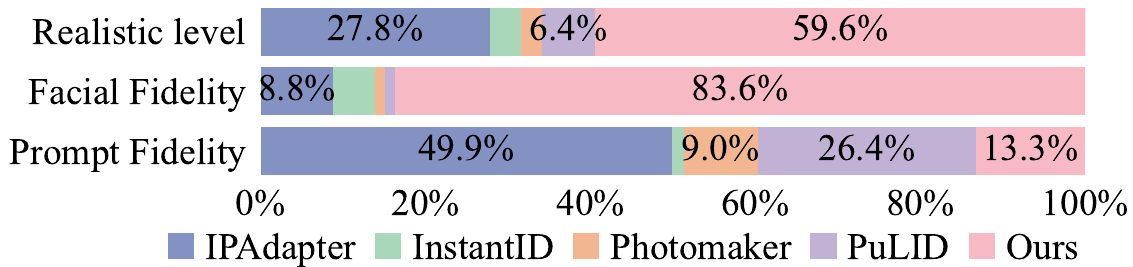}
\caption{Manual evaluation.}
\label{fig:manual}
\end{figure}

Each model generated 100 facial images, which were then mixed together. We randomly invited six graduate students to evaluate their preferences regarding image realism, facial similarity, and text fidelity. As shown in Fig.~\ref{fig:manual}, Face-MakeUp excelled in the two key metrics of image realism and facial similarity. This demonstrates its outstanding performance in visual realism and target identity restoration. However, in terms of text fidelity, our method received only 16.5\% of the votes, falling behind IPAdapter and PuLID. This manual evaluation is consistent with the results in Table~\ref{tab:metrics}.

\subsection{Ablation Study}

(1) \textbf{Effectiveness of FaceCaptionHQ-4M}. Since the data used by many other models have not been made publicly available, we train the classic IPAdapter model from scratch on our FaceCaptionHQ-4M. The evaluations on Unsplash-Face are shown in Table~\ref{tab:ipadapter}. It can be observed that the IPAdapter-FaceID trained on FaceCaptionHQ-4M achieves significant improvements in the quality of generated facial images (CLIP-I, DINO, FaceSim, FID, and VLM-score). This demonstrates that FaceCaptionHQ-4M is beneficial for generating high-quality facial images.

\begin{table}[htbp]
    \centering
    \caption{{IPAdpater-FaceID represents the IPAdpater model trained on FaceCaptionHQ-4M; IPAdpater\cite{ye2023ip} represents the original pre-trained model.}}
    
    \resizebox{0.5\textwidth}{!} {%
    \Huge
    \begin{tabular}{c c c c c c}
        \toprule[3pt]
        
        \textbf{} & \textbf{CLIP-I}\(\uparrow\) & \textbf{DINO}\(\uparrow\) & \textbf{FaceSim}\(\uparrow\) & \textbf{FID}\(\downarrow\) & \textbf{VLM-score}\(\uparrow\) \\ 
        \midrule
        \textbf{IPAdapter}\cite{ye2023ip} & 64.9 & 37.6 & 53.2 & 227.0 & 65.3 \\ 
        \textbf{IPAdapter-FaceID} & 66.2 & 48.12 & 56.8 & 158.7 & 71.8 \\ 
        
        \toprule[3pt]
        
    \end{tabular}   
    }
    \label{tab:ipadapter}
\end{table}

    
        
        
        

(2) \textbf{Effectiveness of Facial Pose}: In the inference, we eliminate facial features one by one. The results on Unsplash-Face are shown in Table~\ref{tab:face_makeup}. We observe that removing facial pose information leads to a decrease in the quality of the generated facial imagess. This demonstrates that pose is also beneficial for Face-MakeUp.

\begin{table}[htbp]
    \centering
    \fontsize{9}{15}

    \caption{\textit{Face-MakeUp$^\dagger$} represents inference without the integration of pose information.}
    \resizebox{0.5\textwidth}{!} {%
    \Huge
    \begin{tabular}{c c c c c c}
        \toprule[2pt]
        \textbf{} & \textbf{CLIP-I}\(\uparrow\) & \textbf{DINO}\(\uparrow\) & \textbf{FaceSim}\(\uparrow\) & \textbf{FID}\(\downarrow\) & \textbf{VLM-score}\(\uparrow\) \\ 
        \hline
        \textbf{Face-MakeUp} & 82.1 & 73.2 & 69.2 & 130.1 & 79.6 \\ 
     
        \textbf{Face-MakeUp$^\dagger$} & 72.0 & 50.8 & 31.3 & 168.0 & 82.4 \\ 
        \toprule[2pt]

    \end{tabular}
    }
    \label{tab:face_makeup}
\end{table}

     



\begin{figure}[htbp]
    \centering
     \includegraphics[width=\linewidth, scale=0.28, trim=20px 20px 20px 20px, clip]{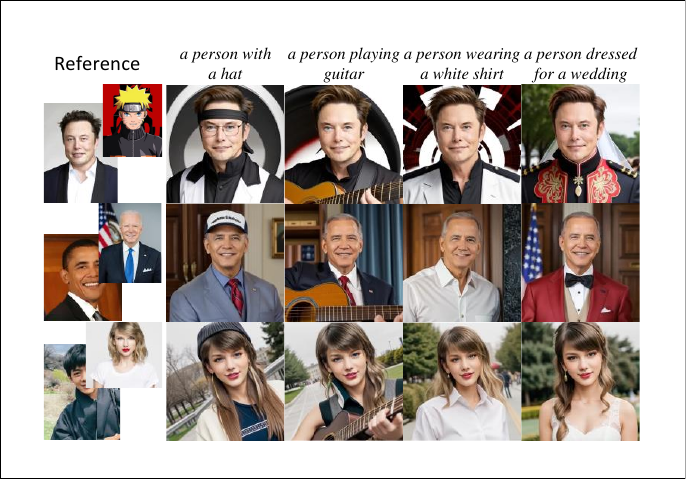}
    \caption{\textbf{Identity mixing.} Face-MakeUp is able to generate the image with a new ID while preserving input identity characteristics}
    \label{fig:identity_mixing}
    \end{figure}

    \vspace{-1em}
\begin{figure}[htbp]
        \centering
        \includegraphics[scale=0.28, trim=2px 2px 2px 2px, clip]{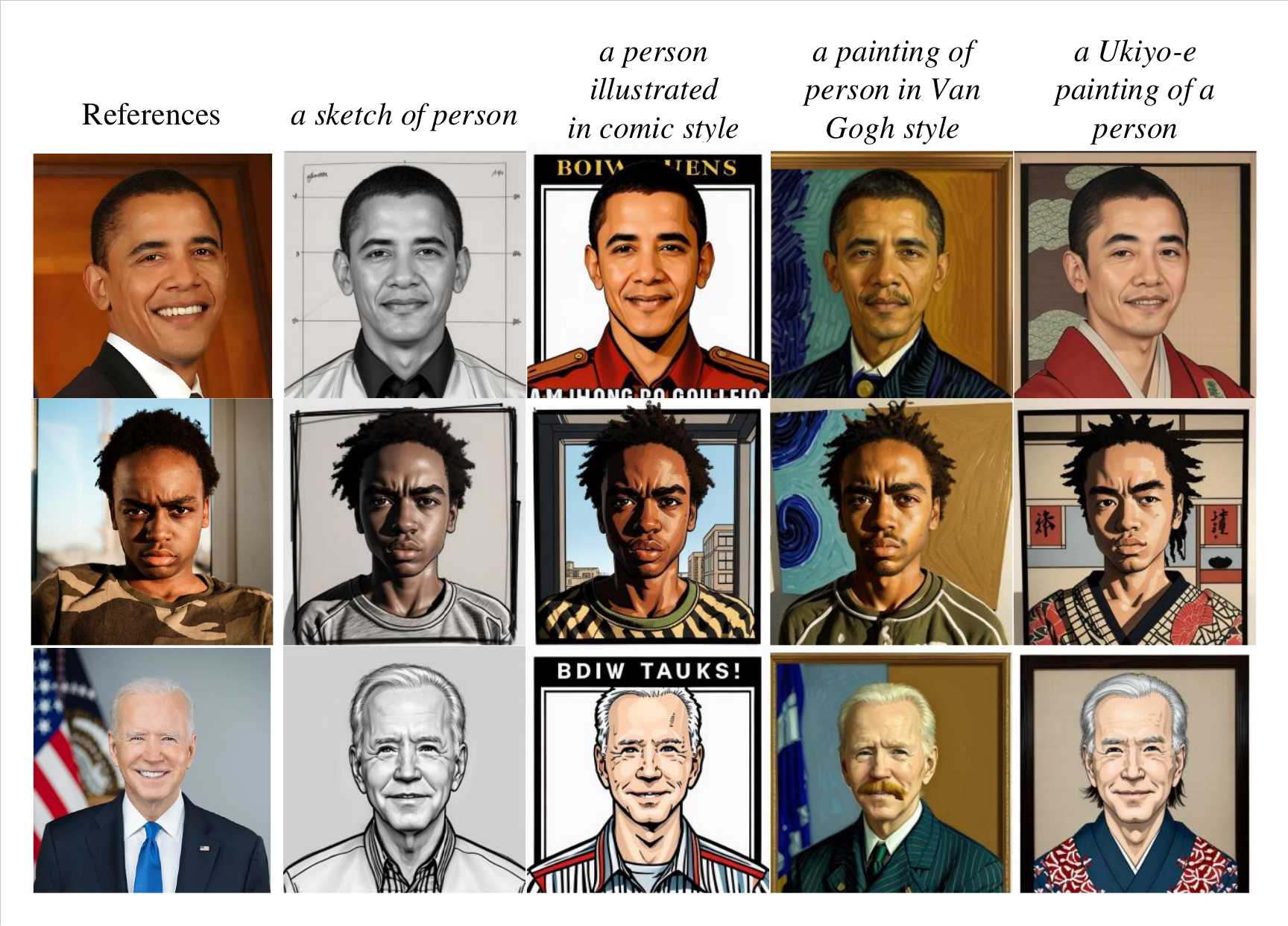}
        \caption{Stylization. Face-MakeUp is able to generate the different styles of facial images using the image and text prompts.}
        \label{fig:stylization}
        \end{figure}

\subsection{Other Applications}

\textbf{Identity mixing.} When users provide images with different identities as input, Face-MakeUp can combine the characteristics from each identity to create a new one. As shown in Fig.~\ref{fig:identity_mixing}, due to the lack of metrics, we rely solely on visual observation, new identity preserves the characteristics of the two input images.

\textbf{Stylization.} Fig.~\ref{fig:stylization} showcases the stylization capabilities of our method. Illustrations demonstrate that Face-MakeUp not only maintains strong face fidelity but also effectively incorporates the style information from the input prompt. This highlights the potential of Face-MakeUp to enable a wide applications.

\section{Conclusion}
General text-to-image models often lack realism and unpredictability when generating facial images. In this work, we have specifically optimized for facial image generation to improve the quality of the generated images. We have presented Face-MakeUp, an efficient method for personalized text-to-image generation, specifically aimed at creating realistic facial images. Experimental results have shown that Face-MakeUp achieves both high-quality and diverse image generation while maintaining strong facial ID fidelity of references. Additionally, we've discovered that our method enables various intriguing applications, such as identity mixing and stylization.

\bibliographystyle{IEEEbib}
\bibliography{icme2025references}

\end{document}